\documentclass[journal]{IEEEtran}
\usepackage{epsfig}
\usepackage{graphicx}
\usepackage{cite}
\usepackage{multirow}
\usepackage{multicol}
\usepackage{amsmath}
\usepackage{amssymb}
\usepackage{latexsym}
\usepackage{tabularx}
\usepackage{booktabs}
\usepackage{color}
\usepackage{makecell}
\usepackage{booktabs}
\usepackage{array}
\usepackage{bm}
\usepackage{diagbox}
\usepackage{url}
\usepackage{hyperref}

\ifCLASSINFOpdf
\else
\fi

\begin{document}

%\bstctlcite{IEEEexample:BSTcontrol}

% correct bad hyphenation here
\hyphenation{op-tical net-works semi-conduc-tor}

%\newcolumntype{Y}{>{\centering\arraybackslash}m{0.7cm}}
%\newcolumntype{K}{>{\centering\arraybackslash}m{2cm}}
%\newcolumntype{s}{>{\centering\arraybackslash}X}
%\newcolumntype{S}{>{\hsize=2.5cm}X}
%\newcolumntype{y}{>{\hsize=1.5cm}X}

\title{Deep Auto-Encoders with Sequential Learning for Multimodal Dimensional Emotion Recognition}

%\author{Michael~Shell,~\IEEEmembership{Member,~IEEE,}
%        John~Doe,~\IEEEmembership{Fellow,~OSA,}
%        and~Jane~Doe,~\IEEEmembership{Life~Fellow,~IEEE}% <-this % stops a space
%\thanks{M. Shell was with the Department
%of Electrical and Computer Engineering, Georgia Institute of Technology, Atlanta,

\author{Dung~Nguyen$^*$,
        Duc~Thanh~Nguyen$^*$,
        Rui~Zeng,
        Thanh~Thi~Nguyen,
        Son~N.~Tran,
        Thin~Nguyen,
        Sridha~Sridharan,~\IEEEmembership{Life~Senior~Member,~IEEE}, 
        and~Clinton~Fookes,~\IEEEmembership{Senior~Member,~IEEE}
\thanks{D. Nguyen, D. T. Nguyen, and T. T. Nguyen are with the School of Information Technology, Deakin University, Australia.}% <-this % stops a space
\thanks{R. Zeng is with the University of Sydney, Australia.}% <-this % stops a space
\thanks{S. N. Tran is with the University of Tasmania, Australia.}
\thanks{T. Nguyen is with the Applied Artificial Intelligence Institute, Deakin University, Australia.}
\thanks{S. Sridharan and C. Fookes are with the Speech, Audio, Image and Video Technology research lab, Queensland University of Technology, Australia.}
\thanks{*\textit{Corresponding author}}
}


\markboth{IEEE Transactions on Multimedia}%
{Shell \MakeLowercase{\textit{et al.}}: Bare Demo of IEEEtran.cls for IEEE Journals}



\maketitle

\begin{abstract}
Multimodal dimensional emotion recognition has drawn a great attention from the affective computing community and numerous schemes have been extensively investigated, making a significant progress in this area. However, several questions still remain unanswered for most of existing approaches including: (i) how to simultaneously learn compact yet representative features from multimodal data, (ii) how to effectively capture complementary features from multimodal streams, and (iii) how to perform all the tasks in an end-to-end manner. To address these challenges, in this paper, we propose a novel deep neural network architecture consisting of a two-stream auto-encoder and a long short term memory for effectively integrating visual and audio signal streams for emotion recognition. To validate the robustness of our proposed architecture, we carry out extensive experiments on the multimodal emotion in the wild dataset: RECOLA. Experimental results show that the proposed method achieves state-of-the-art recognition performance and surpasses existing schemes by a significant margin.
\end{abstract}


% \fi
\IEEEpeerreviewmaketitle

\section{Introduction}
\label{sec:Introduction}

\IEEEPARstart{E}{motion} recognition has become a core research field at the intersection of human communication and artificial intelligence. This research problem is challenging owing to emotions of human beings can be expressed in different forms such as visual, acoustic, and linguistic structures~\cite{Liang:2018:MLR:3242969.3243019}. 

\begin{figure}[t!] 
\centering    
\includegraphics[width=0.5\textwidth]{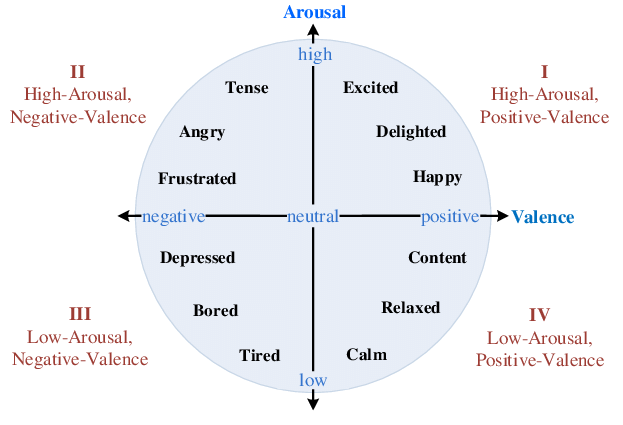}
\caption[A three layer Deep Belief Network]{Two dimensional valence and arousal space (from \url{https://www.pinterest.com.au/pin/354588170647498721/})}
\label{dimensinal emotion}
\end{figure}

As shown in the literature, there are two main conceptualisations of emotions: categorical and dimensional conceptualisation. Categorical approach defines a small set of basic emotions (e.g., happiness, sadness, anger, surprise, fear, and disgust) relying on cross-culture studies that show humans perceive certain basic emotions in similar ways regardless of their culture~\cite{Kollias_2019_2}. Alternatively, dimensional approach represents emotions into a multidimensional space where each dimension captures a fundamental property of the emotions (e.g., appraising human emotional states, behaviours and reactions displayed in real-world settings). These fundamental properties can be accomplished using continuous dimensions in the ``Circumplex Model of Affects'' (CMA)~\cite{8070966} including \textit{valence} (i.e., how positive or negative an emotion is) and \textit{arousal} (i.e., the power of the activation of an emotion). Fig.~\ref{dimensinal emotion} illustrates the CMA. However, this approach is more appropriate to represent subtle changes in emotions, which may not always happen in real-world conditions~\cite{Kollias_2019_2}.

Recently, deep neural networks have been proposed to effectively predict the continuous dimensions of emotions based on multimodal cues such as auditory and visual information~\cite{Kollias_2019_2,8070966,zafeiriou2017aff,kollias2017recognition,7926723,kollias2018aff,kollias2018multi}. These works combine convolutional and recurrent neural networks for feature integration, taking advantages of automatic feature learning in convolutional networks, while encoding temporal dynamics via the sequential layers in recurrent networks~\cite{8070966}. However, feature integration is performed simply by concatenating domain-dependent features extracted from individual modalities. This scheme is straightforward and simple to implement yet may not be able to effectively learn compact and representative multimodal features. To address this issue, we propose a novel deep neural network for multimodal dimensional emotion recognition using a two-stream auto-encoder incorporating with a long short term memory to perform joint learning temporal and compact-representative features in multimodal data. Our architecture can be end-to-end trainable and capable of learning  multimodal representations, achieving state-of-the-art performance on benchmark dataset. Specifically, we make the following contributions,

\begin{itemize}
    \item A two-stream auto-encoder that is able to effectively learn multimodal features from multimodal data for emotion recognition. The constituted auto-encoders ensure compact-representative features to be learnt from individual domains while joint training the auto-encoders captures complementary features from the multiple domains.
    \item A long short term memory that enables sequential learning to encode the long-range contextual and temporal information of multimodal features from data streams. 
    \item Extensive experiments on the multimodal emotion in the wild dataset: RECOLA. In our experiments, ablation studies on various aspects of the proposed architecture are conducted. In addition, important baselines including dimensional facial emotion recognition and dimensional speech emotion recognition, and other existing methods are thoroughly evaluated and analysed.
\end{itemize}

The remainder of this paper is organised as follows: Section~\ref{sec:RelatedWork} briefly reviews related work; Section~\ref{sec:ProposedMethod} describes our proposed method and its related aspects such as data pre-processing, network architecture, training; Section~\ref{sec:Experiments} presents experiments and results; and Section~\ref{sec:Conclusion} concludes our paper with remarks.

\section{Related Work}
\label{sec:RelatedWork}

\noindent Emotion recognition has been a well-studied research problem for several decades and numerous approaches have been proposed. In this section, we limit our review to recent multimodal dimensional emotion recognition methods using deep learning techniques such as convolutional neural networks and long short term memory due to their proven robustness and effectiveness in many applications.

\subsection{Multimodal emotion recognition}
 
\noindent Inspired by the capability of automatic feature learning in deep learning frameworks, Zhang et al.~\cite{7956190} proposed a hybrid deep learning system constructed by a convolutional neural network (CNN), a three-dimensional CNN (3DCNN), and a deep belief network (DBN) to learn audio-visual features for emotion recognition. In this work, the CNN was pre-trained on the large-scale ImageNet database~\cite{ILSVRC15} and used to learn audio features from speech signals. To capture the temporal information from video data, the 3DCNN model in~\cite{Tran:2015:LSF:2919332.2919929} was adapted and fine-tuned on contiguous video frames. The learnt audio and visual features were subsequently fused into the DBN to generate audio-visual features that were finally fed to a linear SVM for emotion recognition. 

To classify spontaneous multimodal emotional expressions, Barros et al.~\cite{7363421} proposed a so-called cross channel convolutional neural network (CC-CNN) to learn generic and specific features of emotions based on body movements and facial expressions. These features were further passed into cross-convolution channels to build cross-modal representations. Motivated by human perception in emotional expression, Barros and Wermter~\cite{Barros:2016:DCE:3040948.3040955} developed a perception representation model capturing the correlation between different modalities. In this work, auditory and visual stimuli were firstly fused using the CC-CNN originally introduced in~\cite{7363421} to achieve a multimodal perception representation. A self-organising layer was then applied on top of the CC-CNN to further separate the perceived expression representations.

\begin{figure*}[t!] 
\centering    
\includegraphics[width=0.9\textwidth]{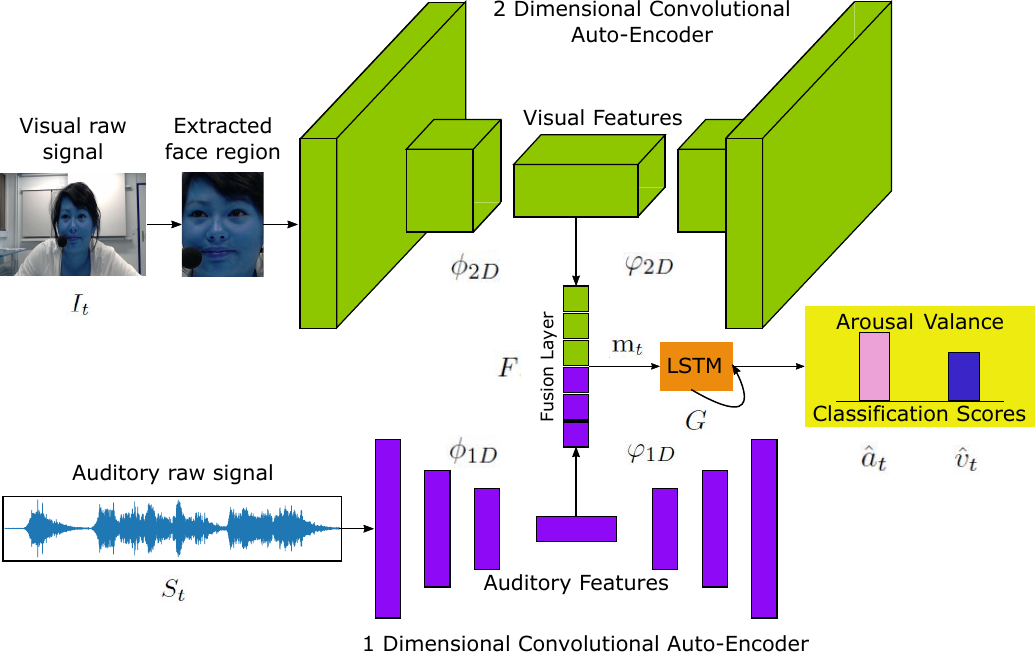}
\caption{Our proposed network architecture for multimodal dimensional emotion recognition.}
\label{fig:NetworkArchitecture}
\end{figure*}

Recently, Zheng et al.~\cite{8283814} proposed a multimodal framework including two Restricted Boltzmann Machines (RBMs) to capture eye movements and EEG signal. The RBMs were unfolded into a bimodal deep auto-encoder (BDAE) \cite{Ngiam:2011:MDL:3104482.3104569} to extract shared representations of the two modalities, which were finally fed to a linear SVM for emotion classification. 

Conventionally, emotion recognition approaches classify person-independent emotions directly from observed data or determine the decreasing/increasing intensity of person-dependent emotions relatively by comparing video segments. However, Liang et al.~\cite{Liang:2018:MLR:3242969.3243019} proposed to combine both approaches for emotion recognition from audio-visual data. In this work, the emotion recognition task was divided into three subtasks including local relative emotion intensities ranking, global relative emotion intensities ranking, and incorporation of emotion predictions from observed multimodal expressions and relative emotion ranks from local-global rankings for emotion recognition.

\subsection{Sequential Learning}

\noindent When temporal information is available, sequential learning can be applied to improve the accuracy of emotion recognition. Long Short Term Memory (LSTM) is often used for sequential learning due to its capability of modelling human memory~\cite{DBLP:journals/corr/KalchbrennerDG15,10.1162/neco.1997.9.8.1735,quteprints115379}. 

Technically, LSTMs are recurrent neural networks (RNNs) integrated with some special gating architectures to control the access to memory cells~\cite{10.1162/neco.1997.9.8.1735}. These gates can also be used to prevent modifying the contents in the cells. LSTMs are, therefore, able to encode much longer range of patterns and propagate errors better than original recurrent neural networks~\cite{NGUYEN201833}. Apart from being able to control the access to the contents in memory cells, the gates can also learn to target on specific parts of input sequences and refuse other parts. This feature allows LSTMs to be able to capture temporal information in sequential patterns.

Inspired by the advantages of LSTMs, many LSTM-based techniques have been developed for human emotion understanding from streaming data. For instance, Chen and Jin~\cite{Chen:2015:MDE:2808196.2811638} and W\"{o}ollmer et al.~\cite{WoLlmer:2013:LCE:2438109.2438270} proposed LSTM-based networks for emotion recognition from data streams. Pei et al.~\cite{7344573} introduced a deep bidirectional LSTM-RNN, which was capable of modelling nonlinear relations in long-term history to handle both multimodal and unimodal emotion recognition tasks. In~\cite{8070966,kollias2019exploiting}, the authors developed an affective recognition system consisting of several deep neural networks to learn features at discrete times and an LSTM to model the temporal evolution of features overtime. 

%In other work, Tzirakis et al.~\cite{8070966} introduced an end-to-end multimodal spontaneous recognition architecture from visual and speech streams. Initially, audio features were extracted from speech signal by a 1-D CNN, while visual features were obtained using a ResNet50 model \cite{7780459}. These features were then concatenated before being fed into a LSTM introduced in \cite{10.1162/neco.1997.9.8.1735} to capture the affective state of individuals.

Despite recent promising achievements, current developments lack of ability to learn compact and representative features on individual domains and effectively learn multimodal features from multimodal data streams. To overcome these limitations, we propose to learn compact and representative features from individual domains using auto-encoders and fuse those domain-dependent features into multimodal features, integrated with temporal information using LSTM.

\section{Proposed Method}
\label{sec:ProposedMethod}

\noindent In this section, we present an end-to-end system for recognition of dimensional emotion from multimodal data including visual and audio data streams.

\subsection{Data Pre-processing}
\label{sec:Preprocessing}

\noindent Input of our system is a pair of video and audio streams. For the video stream, we apply the Single Stage Headless (SSH) detector proposed in~\cite{najibi2017ssh} to detect human faces in video frames. After that we resize the cropped faces to $96 \times 96$. The colour intensities in the cropped images are then normalised to $[-1,1]$.

For the audio stream, we segment the raw waveform signals of the stream into 0.2s long sequences after normalising the time-sequences in order to obtain zero mean and unit variance. The normalisation aims at taking into account the variation in the loudness among different speakers. For a given input stream sampled at 16 kHz, each 0.2s long sequence consists of 5 audio frames, each of which takes 0.04s and is represented by a 640-dimensional vector. Note that each audio frame corresponds to a video frame in the input video stream.

\subsection{Network Architecture}
\label{sec:Network_Architecture}

\noindent The proposed network architecture is illustrated in Fig.~\ref{fig:NetworkArchitecture}. Our architecture consists of two network branches, a 2D convolutional auto-encoder (2DConv-AE) and a 1D convolutional auto-encoder (1DConv-AE), and a long short term memory (LSTM). Each network branch takes its corresponding input, e.g., the 2DConv-AE receives input as an image while the 1DConv-AE receives input as a 0.04s audio frame. The latent layers of these branches are fused into a multimodal representation, which is then fed to the LSTM for prediction of two dimensional emotion scores: arousal and valence.

Given an input video stream (of a speaker) including an image sequence and audio sequence, each image frame in the sequence is passed into the SSH detector~\cite{najibi2017ssh} to detect the speaker's face. The face image is then fed to the 2DConv-AE to learn visual features. Simultaneously, the corresponding audio frame is passed to the 1DConv-AE to learn audio features. The features extracted from the latent layers of these auto-encoders are compact yet representative for their individual domains. Those features are then combined via a fusion layer before being fed to the LSTM for sequential learning of the features (for every 0.2s long sequence) from input streams. The reason for the LSTM is to model the temporal variation of the audio and visual features that provides the contextual information of the input data.

The combination of auto-encoders and LSTM ensures that the learnt representations are compact, rich and complementary and thus make the architecture optimal and robust towards the recognition task from multimodal data streams.

\begin{figure*}
\centering    
\includegraphics[width=0.75\textwidth]{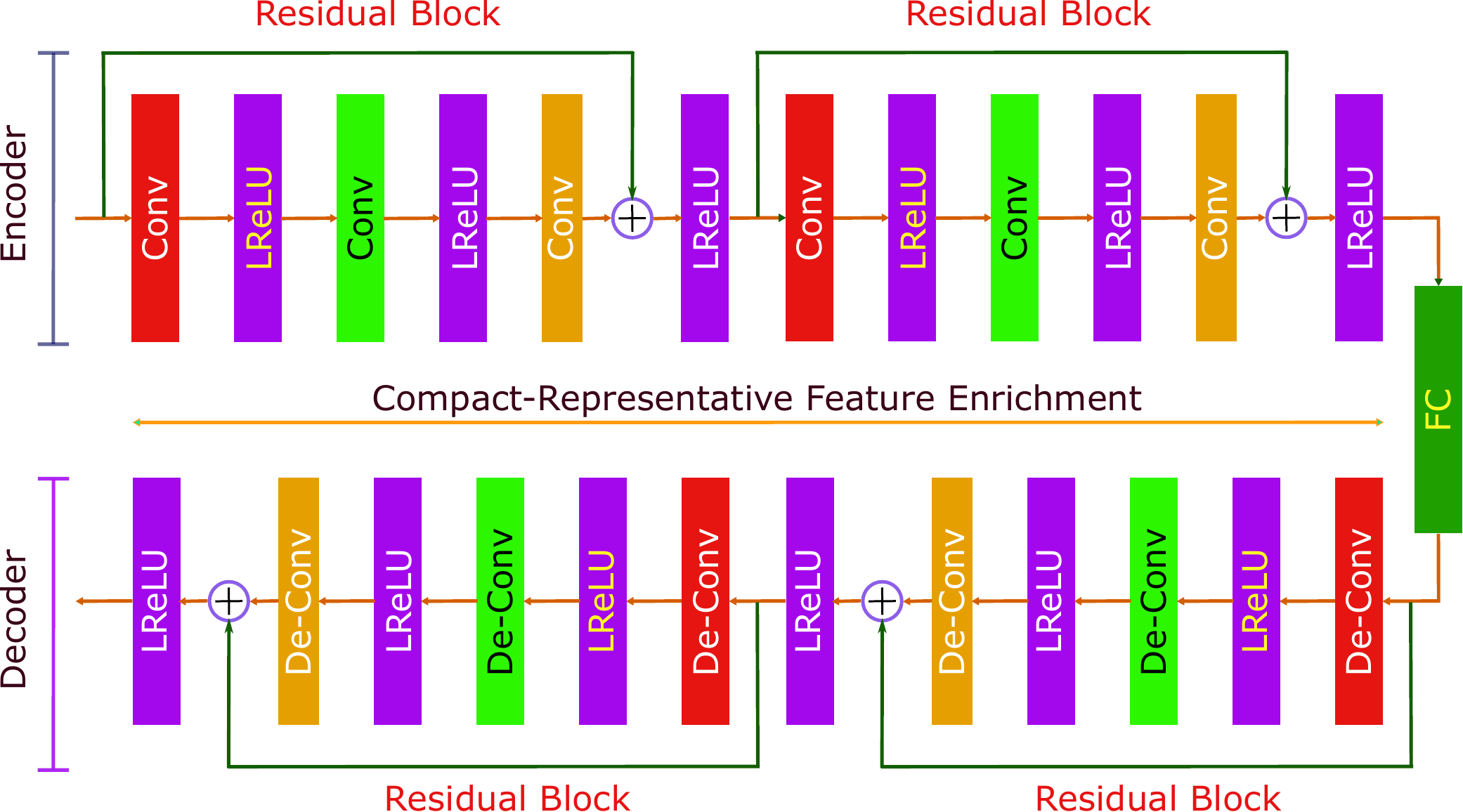}
\caption{Architecture of the 2DConv-AE.}
\label{fig:2DConv-AE}
\end{figure*}

\newcolumntype{G}{>{\centering\columncolor{gray!20!white}}p{0.2\textwidth}}
\newcolumntype{C}{>{\centering\arraybackslash}m{0.078\textwidth}}
\newcolumntype{D}{>{\centering\arraybackslash}m{0.05\textwidth}}
\newcolumntype{K}{>{\raggedleft}m{0.047\textwidth}}
\newcolumntype{E}{>{\centering}p{0.12\textwidth}}

\begin{table}
\begin{center}
\caption{Detailed description of the 2DConv-AE. Conv2D: 2D convolutional layer. DeConv2D: 2D deconvolutional layer. FC: Fully-connected layer.}

\begin{tabular}{KDCEC}
\midrule

\textbf{} & \textbf{Layer} & \textbf{Input size} & \textbf{Filter size, Stride, Out channels}& \textbf{Output size}  \\
\midrule
\multirow{4.5}{*}{Encoder} &
Conv2D & $96 \times 96 \times 3$ &  $\begin{bmatrix} 1 \times 1, 1, 8 \\ 3 \times 3, 2, 8 \\ 1 \times 1, 1, 16 \end{bmatrix}$ & $48 \times 48 \times 16$ \\ 
%\midrule
\cmidrule(r){2-5}

& Conv2D & $48 \times 48 \times 16$ & $\begin{bmatrix}  1 \times 1, 1, 16 \\ 3 \times 3, 2, 16 \\
1 \times 1, 1, 32 \end{bmatrix}$ & $24 \times 24 \times 32$ \\
%\cmidrule(r){2-5}
\midrule
%\multicolumn{2}{c}{}&\multicolumn{3}{c}{2048-d fc1, 2-d fc2}  \\
& FC & 18432 & - & 2048 \\
\midrule
%&FC &640&-&1280\\
%\cmidrule(r){2-5}
\multirow{4.5}{*}{Decoder} & DeConv2D & $24 \times 24 \times 32$ & $\begin{bmatrix}  1 \times 1, 1, 16 \\ 3 \times 3, 2, 16 \\ 1 \times 1, 1, 16 \end{bmatrix}$ & $48 \times 48 \times 16$ \\
\cmidrule(r){2-5}
%\midrule

& DeConv2D & $48 \times 48 \times 16$ & $\begin{bmatrix}  1 \times 1, 1, 8 \\ 3 \times 3, 2, 8 \\ 1 \times 1, 1, 3 \end{bmatrix}$ & $96 \times 96 \times 3$ \\
\midrule
\end{tabular}
\label{tab:2DConv-AE}
\end{center}
\end{table}

\subsubsection{2DConv-AE} follows the common practice of 2D auto-encoders, e.g.,~\cite{NIPS2016_6172}. The encoder of the 2DCov-AE is composed of two residual blocks as in the ResNet architecture~\cite{7780459}, then enclosed by a fully-connected layer. These residual blocks play a central role in feature enrichment. The decoder of the 2DCov-AE includes two residual blocks, which are designed by stacking six 2D de-convolutional layers. Leaky ReLU (LReLU) activation function is adopted after each convolutional layer and de-convolutional layer in our design. Table~\ref{tab:2DConv-AE} provides a detailed description of the 2DConv-AE branch while Fig.~\ref{fig:2DConv-AE} visualises its architecture.

\newcolumntype{H}{>{\centering\arraybackslash}m{0.077\textwidth}}
\newcolumntype{B}{>{\centering\arraybackslash}m{0.06\textwidth}}
\newcolumntype{A}{>{\raggedleft}m{0.042\textwidth}}
\newcolumntype{F}{>{\centering}m{0.120\textwidth}}

\begin{table}
\begin{center}
\caption{Detailed description of the 1DConv-AE. Conv1D: 1D convolutional layer, DeConv1D: 1D deconvolutional layer. FC: Fully-connected layer.}

\begin{tabular}{ABHFH}
\midrule
\textbf{} & \textbf{Layer}  & \textbf{Input size}& \textbf{Filter size, Stride, Out channels}  & \textbf{Output size}\\
\midrule

\multirow{6.6}{*}{Encoder} & Conv1D & $1 \times 640 \times 1$ & [$1\times 20$, 1, 40] & $ 1\times 640 \times 40$ \\ \cmidrule(r){2-5}
& Maxpooling & $1\times 640 \times 40$ & [1, 2, 1] & $1\times 320 \times 40$ \\ \cmidrule(r){2-5}
& Conv1D & $1 \times 320 \times 40$ & [$1 \times 40$, 1, 40] & $1 \times 320 \times 40$ \\ \cmidrule(r){2-5}
& Maxpooling  & $1\times 320 \times 40$ & [1, 10, 1] & $1\times 32 \times 40$ \\ \cmidrule(r){2-5}

& FC & 1280 & - & 640 \\
\midrule
& FC & 640 & - & 1280 \\

\cmidrule(r){2-5}
\multirow{2.2}{*}{Decoder} & DeConv1D & $1 \times 320 \times 4$ & [$1 \times 20$, 1, 40] & $1 \times 320 \times 40$ \\ \cmidrule(r){2-5}
& Upsampling & $1 \times 320 \times 40$ & [1, 2, 1] & $1 \times 64 \times 40$ \\ \cmidrule(r){2-5}
& DeConv1D & $1\times 640 \times 40$ & [$1 \times 40$, 1, 1] & $1 \times 640 \times 1$ \\
\midrule
\end{tabular}
\label{tab:1DConv-AE}
 
\end{center}
\end{table}

\begin{figure}
\centering    
\includegraphics[width=0.5\textwidth]{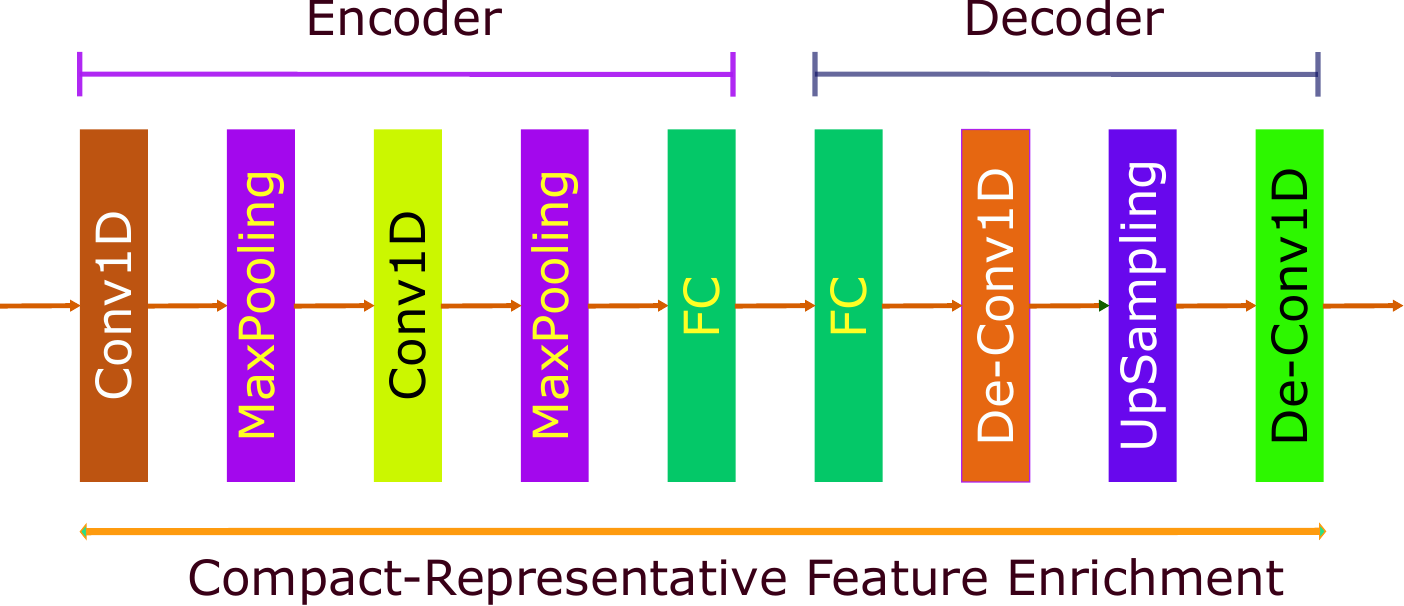}
\caption{Architecture of the 1DConv-AE.}
\label{fig:1DConv-AE}
\end{figure}

\subsubsection{1DConv-AE} realises an 1D auto-encoder applied to audio signal sequences. The encoder of the 1DConv-AE is an 1D convolutional neural network. We adopt the network architecture proposed by Tzirakis et al. \cite{8070966} in our design for the encoder of the 1DConv-AE. Specifically, the 1DConv-AE's encoder includes two 1D convolutional layers, each of which is followed by a max pooling layer. Two fully-connected layers are subsequently attached. The decoder of the 1DConv-AE is formed by stacking one 1D de-convolutional layer, followed by one upsampling layer and then another 1D de-convolutional layer. We summarise the architecture of the 1DConv-AE in Table~\ref{tab:1DConv-AE} and Fig.~\ref{fig:1DConv-AE}.

%\textit{Temporal Convolution:} F = 20 space time finite impulse filters with a 5ms window is used to extract finescale spectral information from the high sampling rate signal.

%\textit{Pooling across time:} The impulse response of each filter is passed through a half-wave rectifier, then downsampled to 8 kHz by pooling each impulse response with a pool size = 2.

%\textit{Temporal Convolution:} M = 40 space time finite impulse filters of 500ms window are used in order to extract more long-term characteristics of the speech and the roughness of the speech signal.

%\textit{Max pooling across channels:} Max-pooling across the channel domain with a pool size of 10 is performed with a focus on reducing the dimensionality of the signal while preserving the necessary statistics of the convolved signal.

\subsubsection{LSTM} has demonstrated a powerful capability of learning long-range contextual information in sequential patterns~\cite{Zhang+2016}. Motivated by this power, we adopt a 2-layer LSTM with 512 cells for each layer to model the temporal and contextual information in multimodal features learnt by the 2DConv-AE and 1DConv-AE. Readers are referred to~\cite{DBLP:series/sci/2012-385} for more detail on LSTM implementation.

\subsection{Joint Learning}
\label{joint_learning}

\noindent Our goal is to learn compact and representative features commonly shared in both visual and auditory domain for prediction of dimensional emotion scores. 

For the ease of presentation, we first introduce important notations used in our method. We denote the encoder and decoder of the 2DConv-AE as $\phi_{2D}$ and $\varphi_{2D}$ respectively. Similarly, the encoder and decoder of the 1DConv-AE are denoted as $\phi_{1D}$ and $\varphi_{1D}$ respectively. Let $F$ denote the fusion layer which concatenates the features learnt by $\phi_{2D}$ and $\phi_{1D}$. The LSTM is represented by $G$, receiving input from the fusion layer $F$. Let $I_t$ denote a facial image obtained by applying the SSH face detector on an input image frame at time step $t$. Let $S_t$ denote the corresponding sound segment of the facial image $I_t$. Given a pair of input $(I_t,S_t)$, let $a_t$ and $v_t$ be the ground truth arousal and valence respectively, and $\hat{a}_t$ and $\hat{v}_t$ be the predicted arousal and valence of the input $(I_t, S_t)$. The training procedure can be described as follows.

Given the input pair $(I_t,S_t)$, to learn compact and representative visual and audio features in individual domains, the auto-encoders 2DConv-AE and 1DConv-AE make use of the 2D and 1D encoders $\phi_{2D}$ and $\phi_{1D}$, and the 2D and 1D decoders $\varphi_{2D}$ and $\varphi_{1D}$, and result in an output pair of reconstructed image frame $\hat{I}_t$ and speech frame $\hat{S}_t$ where,
\begin{equation}
\label{eq:encoder_I}
\hat{I}_t = \varphi_{2D}(\phi_{2D}(I_t)),
\end{equation}
\begin{equation}
\label{eq:encoder_S}
\hat{S}_t = \varphi_{1D}(\phi_{1D}(S_t)).
\end{equation}

The quality of an auto-encoder can be measured via the similarity between an original signal and its reconstructed version after being processed through the auto-encoder. Auto-encoders, thus, can ensure the representative quality of their encoded representations. In this work, we define the losses of our auto-encoders using $\ell_2$-norm as,
\begin{equation}
\label{eq:autoencoderLoss_2D}
\mathcal{L}_{2D} = \sum^{n}_{t=1}\|\hat{I}_t-I_t\|_2^2,
\end{equation}
\begin{equation}
\label{eq:autoencoderLoss_1D}
\mathcal{L}_{1D} = \sum^{n}_{t=1}\|\hat{S}_t-S_t\|_2^2
\end{equation}
where $\hat{I}_t$ and $\hat{S}_t$ is defined in Eq.~(\ref{eq:encoder_I}) and Eq.~(\ref{eq:encoder_S}) respectively, $n$ is the number of samples (i.e., image/speech frames) in a current training batch.

To learn multimodal features, the fusion layer $F$ is used to fuse features extracted from the latent layers of the auto-encoders. Specifically, the multimodal feature vector for the input pair $(I_t,S_t)$ is denoted as $\mathbf{m}_t$ and
defined as,
\begin{equation}
\label{eq:fusion}
\mathbf{m}_t = F(\phi_{2D}(I_t),\phi_{1D}(S_t)) = \phi_{2D}(I_t) \oplus \phi_{1D}(S_t)
\end{equation}
where $\oplus$ represents a concatenating operator.

The multimodal feature vector $\mathbf{m}_t$ is then fed to the LSTM $G$ which estimates the arousal value $\hat{a}_t$ and valence value $\hat{v}_t$ of $\mathbf{m}_t$ based on its precedent observations, i.e., 
\begin{equation}
\label{eq:LSTM}
%(\hat{a}_t,\hat{v}_t) = G(\mathbf{v}_t | \mathbf{v}_{t-1}, ..., \mathbf{v}_{t-k})
(\hat{a}_t,\hat{v}_t) = G(\mathbf{m}_t, \{(\mathbf{m}_{t-i}, \hat{a}_{t-i}, \hat{v}_{t-i})\}_{i=1}^k)
\end{equation}
where $k$ is the number of precedent observations used to determine the arousal and valence value of the observation at time step $t$. 

%In our implementation, $k$ is set to 4, i.e., the previous 4 frames are used given an image/audio frame.

%To learn the visual representation denoted by $\phi_V(\cdot)$ and auditory representation denoted by $\phi_A(\cdot)$, which represent an image $V_t$ and an auditory segment $A_t$ into a dense vector representing an embedding in universal embedding space respectively, while for the visual domain, most existing multimodal emotion recognition systems have utilized a backbone network such as a ResNet101 network, a LSTM or 1-D CNN have been exploited in these systems to model the auditory signal, aiming to learn discriminative representations. Despite of their efficacy of extracting discriminative features from both modalities, we argue that one of the main causes of poor performance of the most existing emotion recognition systems on multimodal emotion recognition task is due to the fact that the visual and the auditory features are still not discriminative enough, and the compact feature representation of these modalities have not been taken into consideration, yet to be of paramount importance for this task.

To measure the quality of arousal and valence prediction, we adopt the Concordance Correlation Coefficient (CCC) proposed in~\cite{10.2307/2532051}. CCC has been widely used in measuring the performance of dimensional emotion recognition systems~\cite{Kollias_2019_2}. It validates the agreement between two time series (e.g., predictions and their corresponding ground truth annotations) by scaling their correlation coefficient with their mean square difference. In this way, predictions that are well correlated with the annotations but shifted in the value are penalised proportionally to their deviations. CCC takes values in the range $[-1, 1]$, where $+1$ denotes perfect concordance and $-1$ indicates perfect discordance. The higher the value of the CCC demonstrates the better the fit between predictions and ground truth annotations, and therefore high values are desired. Applying the CCC, we define the loss for emotion recognition $\mathcal{L}_{Rec}$ as follows,
\begin{equation}
\label{eq:recognitionLoss}
\mathcal{L}_{Rec}=1 - 0.5 \cdot (\rho_{a}+\rho_{v})
\end{equation}
where $\rho_{a}$ and $\rho_{v}$ is the CCC of the arousal and valence respectively, calculated on a current training batch. In particular, we define,
\begin{equation}
\label{eq:rho_a}
\rho_{a} = \frac{2\sigma_{\hat{a},a}}{\sigma^2_{\hat{a}} + \sigma^2_{a} + (\mu_{\hat{a}} - \mu_a)^2}
\end{equation}
\begin{equation}
\label{eq:rho_v}
\rho_{v} = \frac{2\sigma_{\hat{v},v}}{\sigma^2_{\hat{v}} + \sigma^2_{v} + (\mu_{\hat{v}} - \mu_v)^2}
\end{equation}
where, for instance, $\sigma_{\hat{a},a}$ is the covariance of the predictions $\hat{a}$ and ground truth annotations of arousal $a$, $\sigma^2_{\hat{a}}$ and $\sigma^2_{a}$ is the variance of $\hat{a}$ and $a$ respectively, and $\mu_{\hat{a}}$ and $\mu_a$
is the mean of $\hat{a}$ and $a$ respectively. Note that those statistics are calculated on a current training batch. Similar interpretations can be applied to $\rho_v$.

Finally, the loss of entire network is defined as,
\begin{equation}
\label{eq:finalLoss}
\mathcal{L} = \alpha\mathcal{L}_{2D} + \beta\mathcal{L}_{1D} + \gamma\mathcal{L}_{Rec}
\end{equation}
where $\mathcal{L}_{2D}$, $\mathcal{L}_{1D}$, and $\mathcal{L}_{Rec}$ is presented in Eq.~(\ref{eq:autoencoderLoss_2D}), Eq.~(\ref{eq:autoencoderLoss_1D}), and Eq.~(\ref{eq:recognitionLoss}), $\alpha$, $\beta$, and $\gamma$ are weights used to control the influence of sub networks and set empirically.

As shown in Eq.~(\ref{eq:finalLoss}), the individual auto-encoders and the LSTM are jointly trained. This scheme makes the features learnt through the entire network compact, representative, and complementary from different domains.

\section{Experiments}
\label{sec:Experiments}

\begin{figure*}[t] 
\centering    
\includegraphics[width=1.0\textwidth]{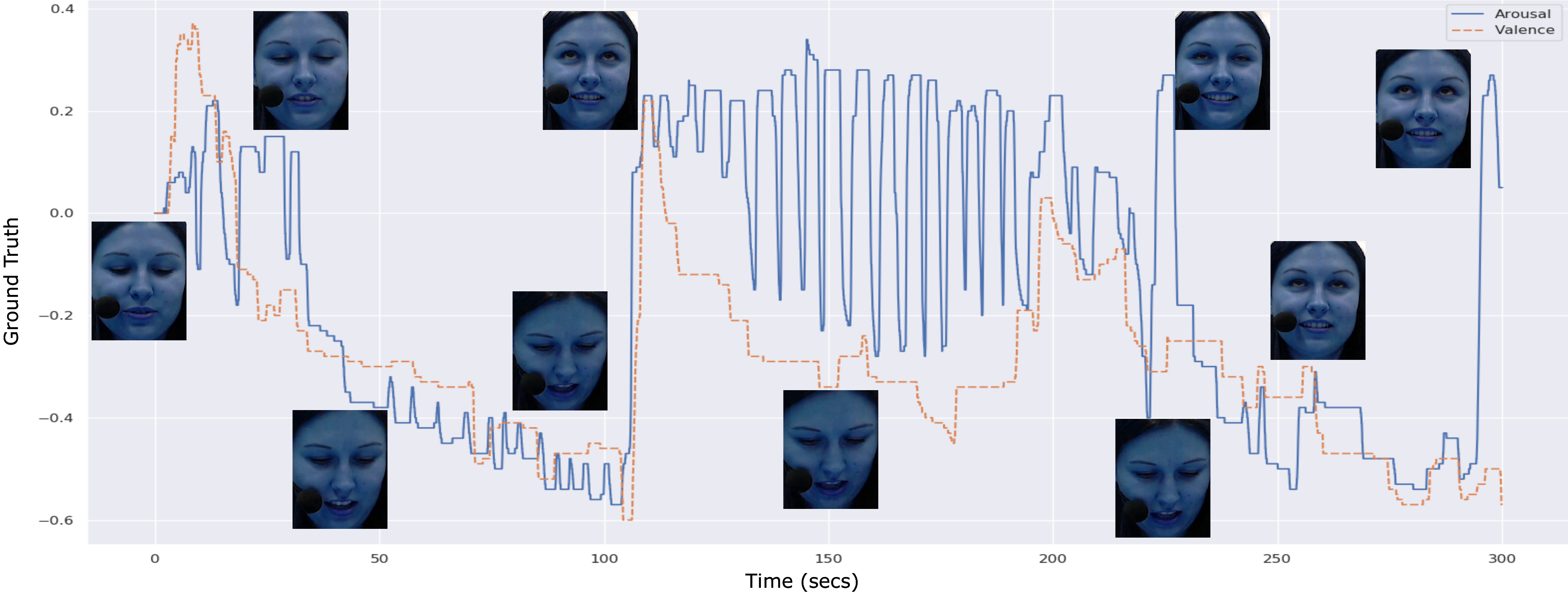}
\caption{Arousal and valance annotations over a part of a video in the RECOLA dataset. Corresponding frames are also illustrated. This figure shows the in-the-wild nature of the emotion data in RECOLA dataset (with different emotional states, rapid emotional changes, occlusions).}
\label{fig:Dataset}
\end{figure*}

\subsection{Dataset} 
\label{sec:dataset}
\noindent We conducted our experiments on the REmote COLlaborative and Affective (RECOLA) dataset introduced by Ringeval et al.~\cite{6553805_1}. This dataset consists of spontaneous and natural emotions represented by continuous values of arousal and valence. The dataset has four modalities including electro-cardiogram, electro-dermal activity, audio, and visual modality. There are 46 French speaking subjects involved in recordings of 9.5 hours in total. The recordings are labelled for every 5 minutes by three male and three female French-speaking annotators. The dataset is balanced among various factors including mother tongue, age, and gender. The dataset includes a training set with 16 subjects and a validation set with 15 subjects. Each subject is associated with a recording (including visual and audio signal). Each recording consists of 7,500 frames for each visual and audio channel. Fig.~\ref{fig:Dataset} illustrates an example in the RECOLA dataset.

\subsection{Implementation Details} 
\label{sec:implementation}
\noindent Our emotion recognition system receives input as a multimodal data stream including an image channel and a speech channel. The data stream of each channel is segmented into frames which are then passed to the proposed network architecture for processing. 

Given a pair $(I_t,S_t)$ including an image frame $I_t$ and its corresponding speech frame $S_t$ at some time step $t$, the image frame $I_t$ is passed to the visual network branch (2DConv-AE) to extract 2,048 visual features via $\phi_{2D}(I_t)$. Similarly, the speech frame $S_t$ is forwarded to the speech network branch (1DConv-AE) to extract 1,280 auditory features via $\phi_{1D}(S_t)$. These output features are concatenated to form a 3,328 dimensional multimodal representation $\mathbf{m}_t$ as defined in Eq.~(\ref{eq:fusion}). This representation is fed to the 2-layer LSTM (with 512 cells per layer) to extract long-range contextual information from the data stream. The output of the LSTM is finally attached to a fully-connected layer to predict the arousal and valence for the input data at time step $t$.

%The result of the LSTM is finally passed to a three fully-connected layer structure including one input layer (from the LSTM's output), one hidden layer with 32 hidden nodes, and one 2-node output layer containing the predicted values for arousal and valence. 

To predict the arousal and valence for the input pair $(I_t,S_t)$, the LSTM takes into account the last four time steps of $t$, i.e., $k$ in Eq.~(\ref{eq:LSTM}) is set to 4.

%Because more advanced methods such as mid-level fusion and late fusion have shown better performance, however they  may require special optimisation and/or parameter tuning. We demonstrate that with the exploitation of simplest early fusion method (e.g., simply done by concatenation of the different input vectors) throughout all fusion based experiments, our proposed system still yields superior performance. Moreover, such usage is to keep the experiments for this work simple and reproducible.

\begin{figure} 
\centering    
\includegraphics[width=0.5\textwidth]{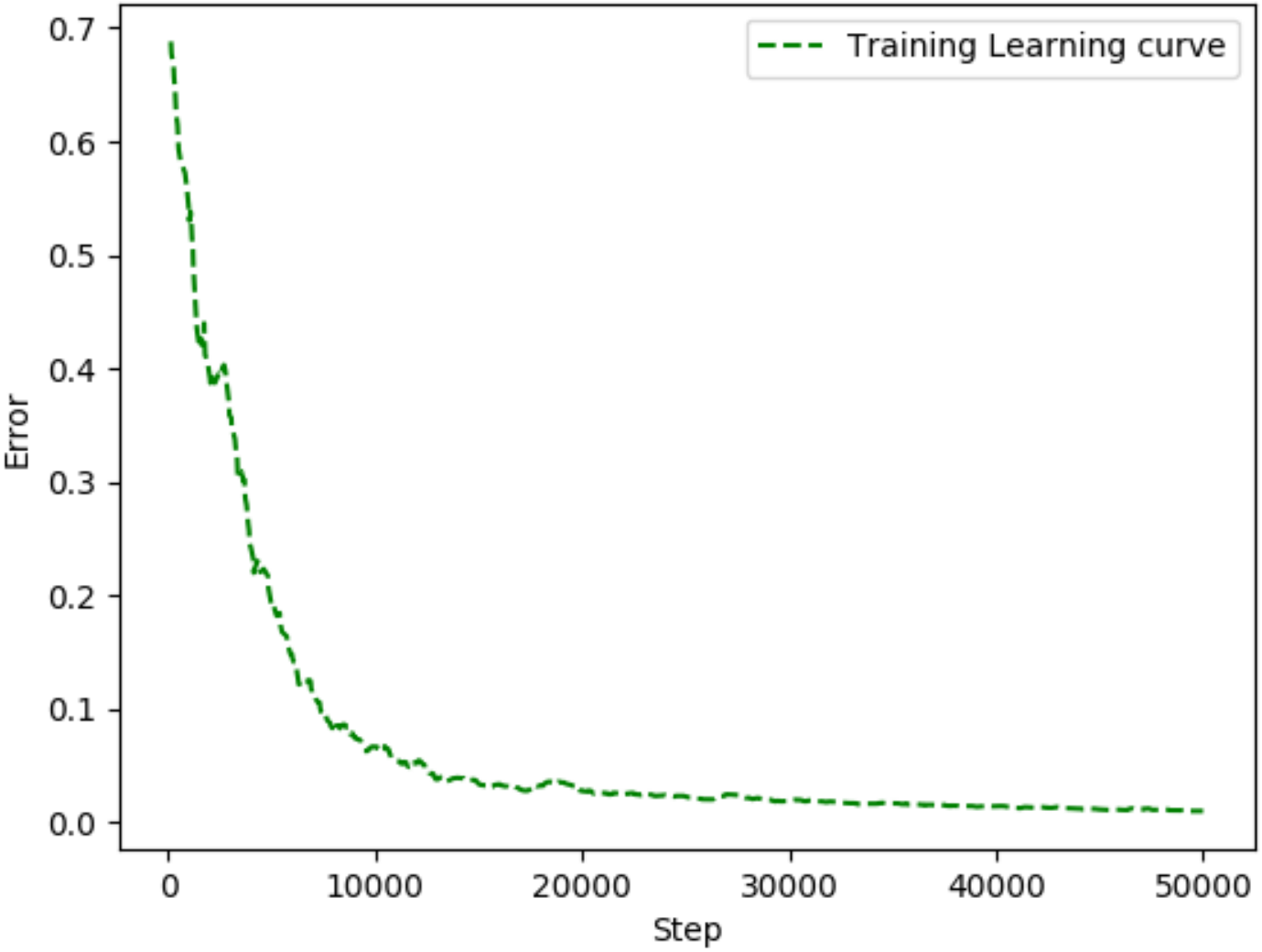}
\caption{Learning curve of our model.}
\label{fig:learningcurve}
\end{figure}

Our proposed architecture was trained end-to-end via optimising the joint loss $\mathcal{L}$ defined in Eq.~(\ref{eq:finalLoss}) on the training set of the RECOLA database. In our implementation, we set the parameters in Eq.~(\ref{eq:finalLoss}) as, $\alpha=1$, $\beta=1$, and $\gamma=0.01$. Adam optimiser (with default values) was adopted. Mini-batch size was set to 32 and learning rate was set to 0.0001. The number of training steps was set to 50,000. All experiments in this paper were implemented in TensorFlow and conducted on 10 computing nodes: 3780 $\times$ 64-bit Intel Xeon Cores. Our proposed model was trained within 45 hours and required 10,485,760KB of memory usage. Fig.~\ref{fig:learningcurve} shows the learning curve of our model.

%We divided each 0.2s input audio sequence into non-overlapping samples of 0.004s. Note that each audio sample is associated with an image frame in the corresponding input video sequence. 

\subsection{Evaluation Protocol}
%\noindent The CCC values defined in Eq.~(\ref{eq:rho_a}), and Eq.~(\ref{eq:rho_v}) are used as a metric to evaluate our model's performance on arousal and valence prediction. This metric has been used widely as the evaluation criterion in affective computing studies and in recent challenges~\cite{8913483}.

%evaluation of continuous emotion recognition , We evaluate our model's prediction performance using Root Mean Square Error (RMSE), which is defined as follows,

\noindent As commonly used in evaluation of dimensional emotion recognition~\cite{Kollias_2019_2}, we measure the Root Mean Square Error (RMSE) of predicted dimensional emotion scores against the ground truth values as follow,
\begin{equation}
\label{eq:RMSE_av}
E_{av}=\sqrt{\frac{1}{N}\sum^N_{i=1}(\hat{a}_i-a_i)^2 + (\hat{v}_i-v_i)^2}
\end{equation}
where $\hat{a}$ and $a$ is the predicted arousal and its ground truth value respectively, and $\hat{v}$ and $v$ is the predicted valence and its ground truth value respectively, $N$ is the total number of frames in an input sequence.

To further investigate the prediction performance in detail, we also calculate the RMSE on each individual emotion dimension as,
\begin{equation}
\label{eq:RMSE_a}
E_{a}=\sqrt{\frac{1}{N}\sum^N_{i=1}(\hat{a}_i-a_i)^2},
\end{equation}
\begin{equation}
\label{eq:RMSE_v}
E_{v}=\sqrt{\frac{1}{N}\sum^N_{i=1}(\hat{v}_i-v_i)^2}
\end{equation}

The RMSE gives us a rough indication of how the derived emotion model is behaving, providing a simple comparative evaluation metric. Small values of RMSE are desired.

\subsection{Evaluations and Comparisons}
\label{sec:evaluations}
\noindent Since there are several technical contributions proposed in this work, e.g., auto-encoders for learning multimodal features, LSTM for sequential learning of multimodal data streams, etc., we evaluate each of these contributions in our experiments. In addition, for each dimension in the evaluations, we also compare our method with existing related works.

\subsubsection{Multimodal vs Unimodal}
We first evaluate emotion recognition using multimodal and unimodal approach. Our proposed multimodal architecture combines two unimodal branches, each of which focuses on a single domain (visual or audio domain). To make a comprehensive comparison between muiltimodal and unimodal approaches, we respectively disabled one of the two branches in our architecture to achieve unimodal architectures. For instance, the unimodal architecture for the visual part is denoted as ``2DConv-AE-LSTM'' and obtained by taking off the speech network branch (see Fig.~\ref{fig:NetworkArchitecture}) while keeping the LSTM which receives input from only the latent layer of the 2DConv-AE. Similar action was applied to the audio part to create ``1DConv-AE-LSTM.'' 

Table~\ref{tab:multimodal_vs_unimodal} compares unimodal and multimodal approaches. As shown in the table, there is an inconsistency in the performance of the unimodal architectures. Specifically, the audio architecture (1DConv-AE-LSTM) outperforms the visual one (2DConv-AE-LSTM) in prediction of arousal while the visual architecture shows better performance than its counterpart in prediction of valence. Compared with these sub-models, our multimodal architecture (``2D1DConv-AE-LSTM'') combining both the unimodal architectures shows superior performance on prediction of both emotion dimensions. 

We also compared our multimodal architecture with other two unimodal architectures proposed in~\cite{8070966}. Specifically, the authors in~\cite{8070966} investigated the combination of CNNs and LSTM on visual and audio domain separately. As shown in Table~\ref{tab:multimodal_vs_unimodal}, our multimodal architecture outperforms all other unimodal ones on both arousal and valence prediction.

%In particular, our method significantly outperforms both its sub-models in prediction of valence while achieving 

%We have also found a discrepancy in 

%In this experiment, the unimodal architecture for the video and audio part is denoted as ``2DConv-AE-LSTM'' and ``1DConv-AE-LSTM.'' As shown in Table~\ref{tab:multimodal_vs_unimodal},  

%\begin{table}
%\caption{Comparison of multimodal and unimodal architectures.}
%\centering
%\begin{tabular}{c||c|c|c|c}
%\hline
%\multirow{2}{*}{\textbf{Architecture}} & %\multicolumn{2}{c|}{\textbf{CCC}} & %\multicolumn{2}{c}{\textbf{MSE}} \\
%\cline{2-5}
%& \textbf{Arousal} & \textbf{Valence} & \textbf{Arousal} & %\textbf{Valence} \\
%\hline
%2DConv-AE-LSTM & -0.010 & 0.036 & 0.289 & 0.046 \\
%\hline
%1DConv-AE-LSTM & 0.185 & -0.003 & 0.243 & 0.056 \\
%\hline
%Our multimodal & 0.102 & 0.155 & 0.215 & 0.035 \\
%\hline
%\end{tabular}
%\label{tab:multimodal_vs_unimodal}
%\end{table}

\begin{table}
\caption{Comparison of multimodal and unimodal architectures. Best performances are highlighted.}
\centering
\begin{tabular}{|l||c|c|c|}
\hline
\backslashbox{\textbf{Architecture}}{\textbf{Metric}}
& $E_a$ & $E_v$ & $E_{av}$ \\
\hline
2DConv-AE-LSTM & 0.538 & 0.214 & 0.579 \\
\hline
1DConv-AE-LSTM & 0.493 & 0.237 & 0.547 \\
\hline
2D model in~\cite{8070966} & 0.476 & 0.192 & 0.514 \\
\hline
1D model in~\cite{8070966} & 0.517 & 0.251 & 0.574 \\
\hline
Our 2D1DConv-AE-LSTM & \textbf{0.474} & \textbf{0.187} & \textbf{0.510} \\
\hline
\end{tabular}
\label{tab:multimodal_vs_unimodal}
\end{table}

\subsubsection{Feature Learning with Auto-Encoders}
We observed that the auto-encoders (2DConv-AE and 1DConvAE) also boosted up the prediction performance of our architecture. In particular, we compared our architecture with the one proposed in~\cite{8070966}, where only CNNs were used to learn the multimodal features. Note that the same LSTM was used in both the architectures. 

We report the prediction performance of our model and the one in~\cite{8070966} in Table~\ref{tab:autoencoders}. As shown in our experimental results, compared with the network architecture proposed in~\cite{8070966}, our model is slightly inferior in prediction of valence (see $E_v$) but significantly more prominent in prediction of arousal (see $E_a$), leading to a better overall performance.

%This deterioration is probably due to the lack of training data as the auto-encoder contains more much parameters then its encoder and decoder component. 

To further investigate the impact of the auto-encoders on  individual domains, we re-designed the unimodal branches by disabling the decoders in those branches and measured their performances. We observed that the 1DConv-AE improved the accuracy of the audio branch on both arousal and valence prediction. Specifically, the improvement was 4.6\% on $E_a$ (from 0.517 to 0.493) and 5.6\% on $E_v$ (from 0.251 to 0.237), leading to an overall improvement of 4.7\% on $E_{av}$ (from 0.574 to 0.547). In contrast, the 2DConv-AE incurred a decrease of 11.5\% on $E_a$ (from 0.476 to 0.538) and 10.3\% on $E_v$ (from 0.192 to 0.214), and 11.2\% on $E_{av}$ (from 0.514 to 0.579). However, as shown in our experimental results, the incarnation of both the 2DConv-AE and 1DConv-AE compensated the weakness of each individual component and achieved improved overall performance.

\begin{table}
\caption{Evaluation of auto-encoders. Best performances are highlighted.}
\centering
\begin{tabular}{|l||c|c|c|}
\hline
\backslashbox{\textbf{Architecture}}{\textbf{Metric}}
& $E_a$ & $E_v$ & $E_{av}$ \\
\hline
2D1DConv-LSTM~\cite{8070966} & 0.488 & \textbf{0.184} & 0.522 \\
\hline
Our 2D1DConv-AE-LSTM & \textbf{0.474} & 0.187 & \textbf{0.510} \\
\hline
\end{tabular}
\label{tab:autoencoders}
\end{table}

\subsubsection{Sequential Learning with LSTM}
We propose the use of LSTM for learning long-range contextual and temporal information from streaming data. LSTM has also been used widely in dimensional emotion recognition from data streams. For instance, in~\cite{kollias2019exploiting}, Kollias and Zafeiriou proposed to use two LSTMs, each of which for one unimodal data stream (e.g., visual or audio stream). We denote this approach as ``2D1D-2SLSTM.'' Unlike~\cite{kollias2019exploiting}, in our architecture, we use only one LSTM receiving input from a fusion layer and producing predicted values for arousal and valence. Note that auto-encoders were not employed in~\cite{kollias2019exploiting}. Therefore, to better study the effect of using one LSTM vs two stream LSTM, we applied the two stream LSTM in~\cite{kollias2019exploiting} to our architecture, to create a so-called ``2D1DConv-AE-2SLSTM'' variant. 

Table~\ref{tab:LSTM} compares different approaches using LSTM for sequential learning in dimensional emotion recognition. As shown in the table, our architecture achieves the best performance among all models on both emotion dimensions.

\begin{table}
\caption{Evaluation of LSTM. Best performances are highlighted.}
\centering
\begin{tabular}{|l||c|c|c|}
\hline
\backslashbox{\textbf{Architecture}}{\textbf{Metric}}
& $E_a$ & $E_v$ & $E_{av}$ \\
\hline
2D1DConv-2SLSTM~\cite{kollias2019exploiting} & 0.493 & \textbf{0.187} & 0.527 \\
\hline
2D1DConv-AE-2SLSTM & 0.508 & 0.190 & 0.542 \\
\hline
Our 2D1DConv-AE-LSTM & \textbf{0.474} & \textbf{0.187} & \textbf{0.510} \\
\hline
\end{tabular}
\label{tab:LSTM}
\end{table}

\subsubsection{Ablation Study} In this experiment, we study the effect of hidden nodes in the LSTM. Specifically, we investigate the prediction performance of our architecture with regard to various numbers of hidden nodes in each layer in the LSTM. 

We report these results in Table~\ref{tab:ablationstudy}. In general, there is a fluctuation in the prediction performance when varying the number of hidden nodes in the LSTM. Although the best configuration for prediction of valence is the LSTM with 256 nodes in each hidden layer, our current setting with 512 nodes in each hidden layer shows better performance in prediction of arousal and also achieves the best overall performance.

\begin{table}
\caption{Prediction performance of our architecture when varying the number of hidden nodes used in each layer of the LSTM. Best performances are highlighted.}
\centering
\begin{tabular}{|l||c|c|c|}
\hline
\backslashbox{\textbf{\#Hidden nodes}}{\textbf{Metric}}
& $E_a$ & $E_v$ & $E_{av}$ \\
\hline
32 & 0.509 & 0.190 & 0.543 \\
\hline
64 & 0.540 & 0.195 & 0.574 \\
\hline
128 & 0.502 & 0.205 & 0.542 \\
\hline
256 & 0.496 & \textbf{0.184} & 0.529 \\
\hline
512 (our architecture) & \textbf{0.474} & 0.187 & \textbf{0.510} \\
\hline
\end{tabular}
\label{tab:ablationstudy}
\end{table}

\section{Conclusion}
\label{sec:Conclusion}

\noindent This paper proposes a deep network architecture for end-to-end dimensional emotion recognition from multimodal data streams. The proposed architecture incarnates auto-encoders for learning multimodal features from visual and audio domains, and LSTM for learning contextual and temporal information from streaming data. Our proposed architecture enables learning compact, representative, and complementary features from multimodal data source. We implemented various baseline models and conducted extensive experiments on the benchmark RECOLA dataset. Experimental results confirmed the contributions of our work and the superiority of our proposed architecture over the state-of-the-art.

%show that the multimodal approach improved the overall prediction performance and the proposed architecture outperformed state-of-the-art models.

%\appendix[Proof of t

% or



%\section*{Acknowledgment}


\ifCLASSOPTIONcaptionsoff
  \newpage
\fi



%\bibliographystyle{IEEEtran}

\bibliographystyle{IEEEtran}
\bibliography{IEEEexample}


\begin{IEEEbiography}[{\includegraphics[width=1in,height=1.25in,clip,keepaspectratio]{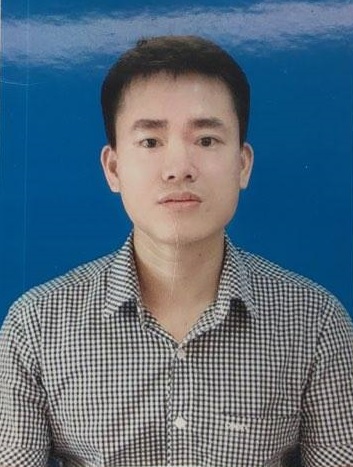}}]{Dung Nguyen}
received the B.Eng. and M.Eng. degrees from the People Securiry Academy and the Vietnam National University – University of Enginerring and Technology in 2006 and 2013,
respectively. He received his PhD degree in the area of multimodal emotion recognition using
deep learning techniques from Queensland University of Technology in Brisbane, Australia in
2019. He is currently working as a research fellow at Deakin University. His current
research interests include computer vision, machine learning, deep learning, image processing,
and affective computing.
\end{IEEEbiography}

\begin{IEEEbiography}[{\includegraphics[width=1in,height=1.25in,clip,keepaspectratio]{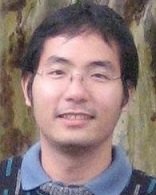}}]{Dr Thanh Nguyen}
was awarded a PhD in Computer Science from the University of Wollongong, Australia in 2012. Currently, he is a lecturer in the School of Information Technology, Deakin University, Australia. His research interests include computer vision and pattern recognition. He has published his work in highly ranked publication venues in Computer Vision and Pattern Recognition such as the Journal of Pattern Recognition, CVPR, ICCV, ECCV. He also has served a technical program committee member for many premium conferences such as CVPR, ICCV, ECCV, AAAI, ICIP, PAKDD and reviewer for the IEEE Trans. Intell. Transp. Syst., the IEEE Trans. Image Process., the IEEE Signal Processing Letters, Image and Vision Computing, Pattern Recognition, Scientific Reports.
\end{IEEEbiography}

\begin{IEEEbiography}[{\includegraphics[width=1in,height=1.25in,clip,keepaspectratio]{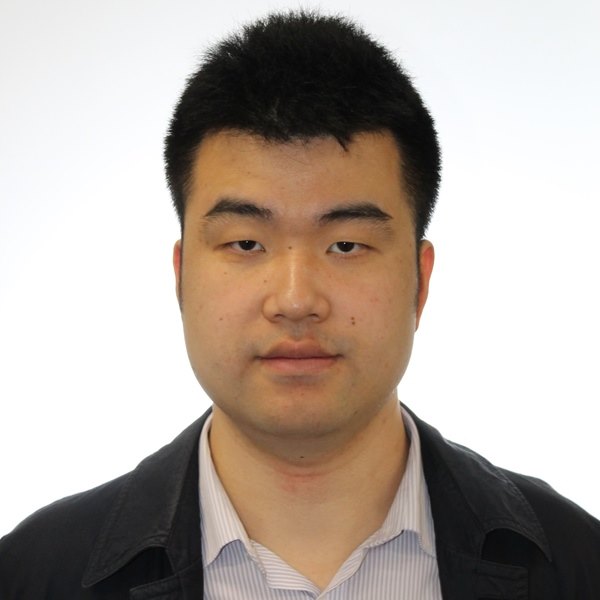}}]{Rui Zeng} received the B.S. degree in Computer Science from Wuhan Institute of Technology in 2012, the M.S. degree in electronic telecommunications in University of Rennes 1, the M.Eng. degree in Computer Science from Southeast University in 2015, and the PhD degree in the area of homography estimation from Queensland University of Technology in 2020. He was a research fellow in Australian Centre for Robotic Vision and eResearch Centre at Monash University. Dr Zeng is currently a postdoctoral research associate in the Brain and Mind Centre at the University of Sydney, Australia. His research interests include multi-view geometry, deep learning, and medical imaging.
\end{IEEEbiography}

\begin{IEEEbiography}[{\includegraphics[width=1in,height=1.25in,clip,keepaspectratio]{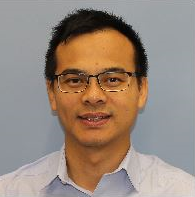}}]{Thanh Thi Nguyen}
was a Visiting Scholar with the Computer Science Department at Stanford University, California, USA in 2015 and the Edge Computing Lab, John A. Paulson School of Engineering and Applied Sciences, Harvard University, Massachusetts, USA in 2019. He received an Alfred Deakin Postdoctoral Research Fellowship in 2016, a European-Pacific Partnership for ICT Expert Exchange Program Award from European Commission in 2018, and an Australia–India Strategic Research Fund Early- and Mid-Career Fellowship Awarded by The Australian Academy of Science in 2020. Dr Nguyen obtained a PhD in Mathematics and Statistics from Monash University, Australia in 2013 and has expertise in various areas, including artificial intelligence, deep learning, deep reinforcement learning, cyber security, IoT, and data science. He is currently a Senior Lecturer in the School of Information Technology, Deakin University, Victoria, Australia.
\end{IEEEbiography}
\begin{IEEEbiography}[{\includegraphics[width=1in,height=1.25in,clip,keepaspectratio]{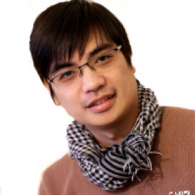}}]{Son N. Tran}
is a Lecturer at the University of Tasmania. Son was a post-doctoral research fellow at the Commonwealth Scientific and Industrial Research Organisation, Australia. He holds a Ph.D. in Computer Science (2016) from City, University of London, an EU Erasmus Mundus joint MSc in Networks and e-Business Computing, University of Reading, and a BSc and MSc in Electronic Engineering from the Hanoi University of Technology. Tran is the recipient of the Erasmus Mundus certificate of achievement 2010 in recognition of an outstanding academic performance. His current research interest is learning and reasoning with visual objects.
\end{IEEEbiography}
%\begin{IEEEbiography}[{\includegraphics[width=1in,height=1.25in,clip,keepaspectratio]{IEEEtran/figure/mohamed.jpeg}}]{Mohamed Abdelrazek}
 %is an Associate Professor of Software Engineering and IoT and founder of Deakin Launchpad for industry innovation at Deakin University. Mohamed has more than 15 years of the software industry, research and teaching experience. Before joining Deakin University in 2015, Mohamed worked as a senior research fellow at Swinburne University of Technology and Swinburne-NICTA software innovation lab (SSIL). Mohamed's research interests include model-driven engineering, adaptive software systems, and engineering of AI software systems.
%\end{IEEEbiography}

\begin{IEEEbiography}[{\includegraphics[width=1in,height=1.25in,clip,keepaspectratio]{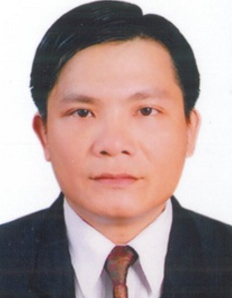}}]{Thin Nguyen} is a Senior Research Fellow with the Applied Artificial Intelligence Institute (A2I2), Deakin University, Australia. He graduated with a PhD in Computer Science from Curtin University, Australia in 2012 in the area of machine learning and social media analytics. His current research topic is strongly inter-disciplinary, bridging large-scale data analytics, pattern recognition, genetics and medicine. His research direction is to utilize artificial intelligence methods to discover functional connections between drugs, genes and diseases. The domain of application emphasizes on personalized medicine, where he has developed personalized annotation-based networks for the prediction of breast cancer relapse. In that line of work, he won the first prize in the DREAM Single Cell Transcriptomics Challenge, organized by IBM Research \& Sage Bionetworks.
\end{IEEEbiography}

\begin{IEEEbiography}[{\includegraphics[width=1in,height=1.25in,clip,keepaspectratio]{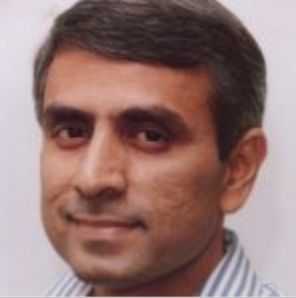}}]{Sridha Sridharan} has a BSc (Electrical Engineering) degree and obtained a MSc (Communication Engineering) degree from the University of Manchester UK and a PhD degree in the area of Signal Processing from University of New South Wales, Australia. He is a Life Senior Member of the Institute of Electrical and Electronic Engineers - IEEE (USA). He is currently with the Queensland University of Technology (QUT) where he is a Professor in the School Electrical Engineering
and Robotics. Professor Sridharan is the Leader of the Research Program in Speech, Audio, Image and Video Technology (SAIVT) at QUT. He has published over 500 papers consisting of publications in journals and in refereed international conferences in the areas of Image and Speech technologies. He has graduated over 80 PhD students at QUT. Prof Sridharan has also received a number of research grants from various funding bodies including commonwealth competitive funding schemes such as the Australian Research Council (ARC) and Cooperative Research Centres (CRC). Several of his research outcomes have been commercialised.
\end{IEEEbiography}

\begin{IEEEbiography}[{\includegraphics[width=1in,height=1.25in,clip,keepaspectratio]{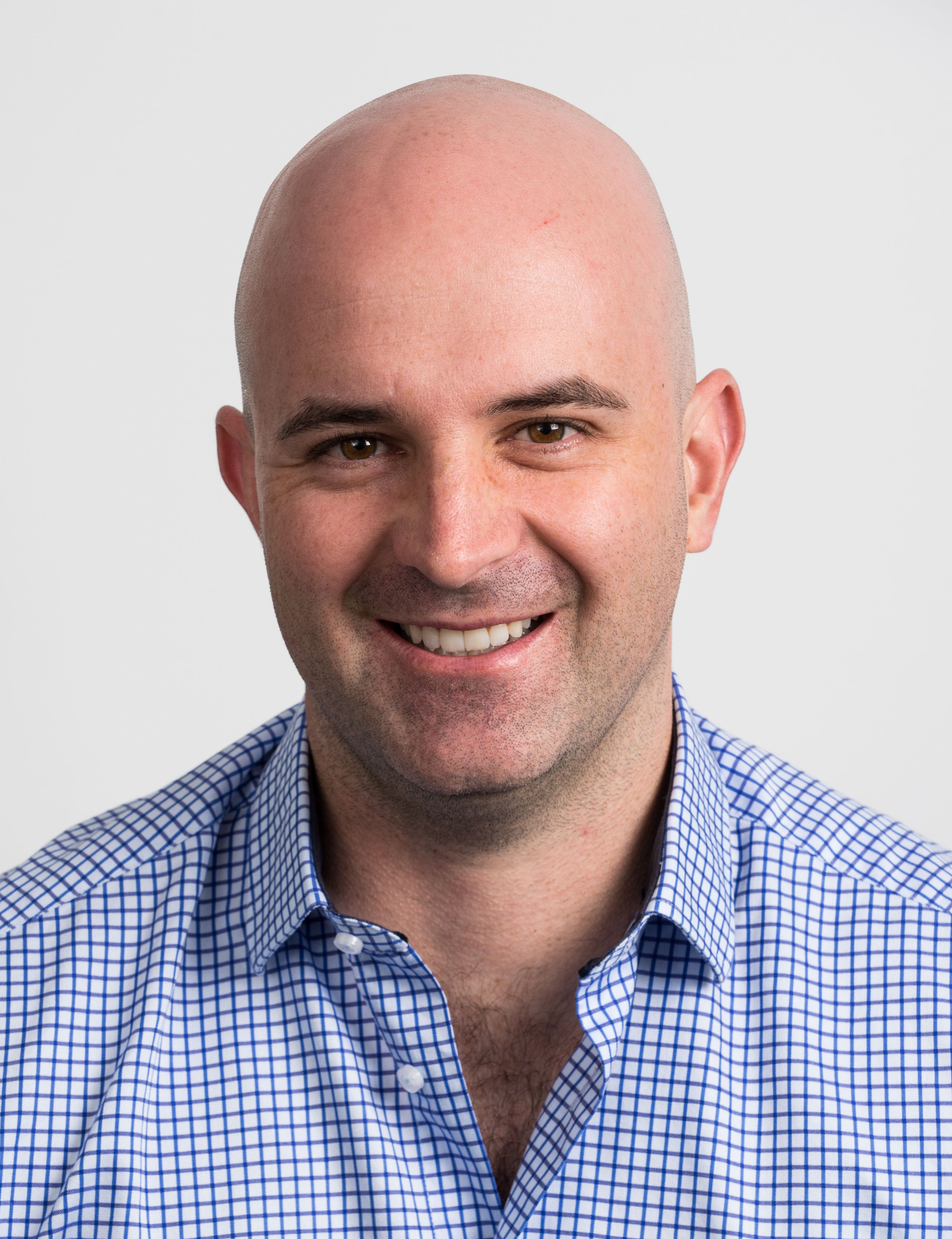}}]{Clinton Fookes} is a Professor in Vision \& Signal Processing at the Queensland University of Technology. He holds a BEng (Aerospace/Avionics), an MBA, and a PhD in computer vision. He actively researches across computer vision, machine learning, signal processing and pattern recognition areas. He serves on the editorial boards for the Pattern Recognition Journal and the IEEE Transactions on Information Forensics \& Security. He is a Senior Member of the IEEE, an Australian Institute of Policy and Science Young Tall Poppy, an Australian Museum Eureka Prize winner, and a Senior Fulbright Scholar.
\end{IEEEbiography}


%\vfill


\end{document}